# A Comparism of the Performance of Supervised and Unsupervised Machine Learning Techniques in evolving Awale/Mancala/Ayo Game Player

Randle, O.A [1] ., Ogunduyile, O.O [2] ., Zuva T [3] Fashola N.A [4]

Tshwane University of Technology [1,2,3] College Campus [4]

Department of Computer Science [1,2] , Department of Computer Engineering [3]

Soshanguve

*ABSTRACT*

*Awale games have become widely recognized across the world, for their innovative strategies and techniques which were used in evolving the agents (player) and have produced interesting results under various conditions. This paper will compare the results of the two major machine learning techniques by reviewing their performance when using minimax, endgame database, a combination of both techniques or other techniques, and will determine which are the best techniques.*

## 1. Introduction

Games are activities of interest to every individual both adults and children. Games are used to learn skills, prepare for tactical activities such as military training and give individuals the ability to compete against each other [1,2]. Computer games are an aspect of machine learning, other aspects of machine learning include robotics, computer vision [3] and machine learning is an aspect of Artificial Intelligence (AI). Computer games include Baganom, Awale, and Chess [4].

African board games have assisted children in counting [5] and thinking intelligently and forecasting. Awale as a game comes from the family of MANCALA and can be referred to by various names such as Ayo, Ayoayo, Awele, Oware [2]. The aim of the game is to capture more seeds than the opponent and win the game.

Awale is a two –person-zero-sum board game consists of 12 pits on two rows called as usual, North and South, with 4 seeds in each pit at the beginning of a game [6]. The rules applied include a player selects all seeds from a non-empty pit on his row and sows them counter-clockwise into each pit excluding the starting pit [6]. If the last seed is sown into a pit on the opponent's row, leaving that pit with 2 or 3 seeds, the player captures the seeds in the pit and seeds in preceding pits on the opponent's row that contain 2 or 3 seeds (this is called the 2-3 capture rule).

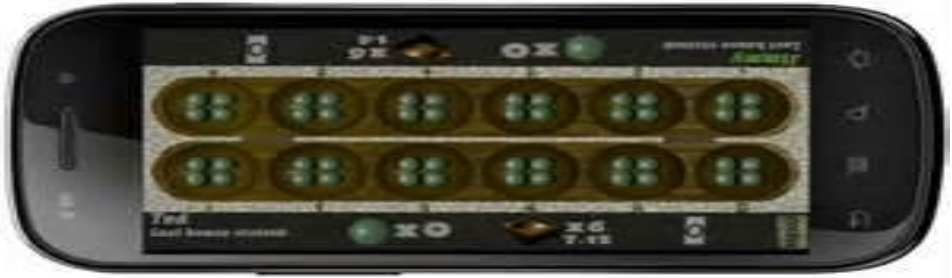

Figure 1. A digital version of Awale Game [7]

A player cannot capture all the seeds on the opponent's row, so he is obliged to make a move that will give his opponent a move and this is called the golden rule. A controversial rule of Awale, yet to be resolved, is when a player cannot move in such a way that he gives his opponent a legal move, then either the game is cancelled or the player that caused this stalemate loses the game no matter his score. The game comes to a conclusion if one of the 3 events occur:

- when a player has captured more than 24 seeds, or
- when both players have captured 24 seeds leading to a draw or
- when fewer seeds circulate endlessly on the board. Case (3) has the following specialisation: if there are fewer seeds on the board that neither player can ever capture, but both players will always have a legal move, the game ends and each player is awarded the seeds on his row.

Machine Learning can be divided into supervised learning and unsupervised, section 2 will discuss supervised machine learning techniques , section 3 will discuss unsupervised machine learning techniques , section 4 compares the results of various techniques which have been used to evolve Awale game player[2,7] to see the performance of both to see what components make or enhance the performance .Various techniques have been implemented to evolve awale player and these techniques can be classified either based on endgames or search technique implemented.

## 2. Supervised Machine learning Techniques(SMLT)

Supervised machine learning is based on the idea of creating a machine that can think or reason outside the box and be able to produce hypothesis or results [8]. All supervised machine learning techniques follow a set of designed principles or models which contain problems, identification of required data[9],pre-processing[10], algorithm selection[11], training[12], and evaluation[7].These supervised techniques can be divided or grouped as logical based algorithms such as decision trees[13], Perceptron based techniques such as single layered perceptron[14], Statistical learning algorithms such as Bayesian Networks[15],Instance based learning and Support vector machines[8].

Supervised machine techniques have been implemented in evolving Awale game player using various techniques such as Case Based Reasoning(CBR)[16], Linear Discriminate Algorithm(LDA)[12], Re-Assisted-Minimax algorithm(RAM)[4], Genetic Algorithm(GA)[17], Co-Evolution (Co-evo)[18]and have produced several results from their performance against the awale shareware and have been sufficiently discussed in previous studies such as[2,7] which has analyzed and investigated the limitations of the aforementioned techniques, and the performance of these techniques are anaysed in table 3.

The best performance for supervised learning has come from Case Based Reasoning(CBR)[16] and the refinement procedure called "casing"[14] .They were able to defeat Awale sufficiently at the grandmaster level or stage. Casing is a combination of case-based reasoning [4] and perceptron learning which acted as the basic move classification algorithm. This method assisted the evolved player by determining the source episodes which are the closest neighborhood to the target episode at the training phase [7]. The similarity, sim ($x_i$ and $y_j$) between two episodes $x_i$ and $y_j$ was calculated using equation 1 which is the product-moment formula for linear correlation coefficient[19].

$$sim(x_i, y_j) = \frac{\sum_{k=1}^{m}(x_{ik} - x_{ai})(y_{jk} - y_{aj})}{\sqrt{\sum_{k=1}^{m}(x_{ik} - x_{ai})^2 \sum_{k=1}^{m}(y_{jk} - y_{aj})^2}}$$

(1)

The evolved player OPON (the name of the evolved player) defeated Awale at all stages but Table 1 shows the performance at the amateur and grandmaster stage/levels. At the grandmaster stage it defeated Awale grandmaster by 25.17 points [14].

Table I. The refinement process (Casing)

| CASING | | | |
|---|---|---|---|
| LEVEL | AVERAGE(MOVES) | SEEDS CAPTURED BY EVOLVED PLAYER(STD) | SEEDS CAPTURED BY AWALE (STD) |
| AMATEUR | 48.33(18.8) | 25.17(0.41) | 14.17(1.60) |
| GRANDMASTER | 41.50(2.74) | 25.50(0.55) | 15.00(1.00) |

The other successful technique is a combination of minimax search and Case based reasoning (CBR) which is designed or based on the concept of using old technique or ideas to solve a new problem. This technique uses a reasoner which assists by remembering the previous problem and the solution which was used to solve the problem [19]. It furthermore combines equation 1 with the minimax search technique. At the testing phase new episode is discovered and its similarities to the source episodes are calculated, where the similarity between ith target episode $x_i = (x_{i1}, x_{i2}, x_{i3},....., x_{im})$ and the source episode $y_j = (y_{j1}, y_{j2}, y_{j3},......, y_{jm})$ of the Jth class is computed. Note - That the target episode with game value $\leq \alpha$ and similarity measure $\geq \beta$ was selected. The similarity was denoted by $\text{Sim}(x_i, y_j)$ between $x_i$ and $y_j$ was calculated using the product-moment formula for the linear correlation coefficient.

In minimax search the value of a leaf is determined by the evaluator and represents the number in proportion to the probability of winning the game. The evaluator can be extended to the minimax function, which determines the value for each player in a node and is formally given in (1) as follows [30,31]:

$$f(n) = \begin{cases} eval(n), & \text{if } n \text{ is a leaf node} \\ \max\{f(c) \mid c \text{ is a child node of } n\}, & \text{if } n \text{ is a max node} \\ \min\{f(c) \mid c \text{ is a child node of } n\}, & \text{if } n \text{ is a min node} \end{cases} \quad (2)$$

The function eval(n) scores the resulting board position at each leaf node n. The standard method of scoring is in terms of a linear polynomial [32]. It has been shown that every game tree algorithm constructs a superposition of a max ($T^+$) and a min($T^-$) solution tree. The equivalent evaluator is the following Stockman equality [33]:

$$f(n) = \begin{cases} \max\{g(T^-) \mid T^- \text{ is a min tree rooted in } n\} \\ \min\{g(T^+) \mid T^+ \text{ is a max tree rooted in } n\} \end{cases} \quad (3)$$

Where the function g is defined by [18]:
$$g(T^+) = \max\{f(c) \mid c \text{ is a terminal in } T^+\}$$
$$g(T^-) = \min\{f(c) \mid c \text{ is a terminal in } T^-\} \quad (4)$$

Conventionally, the basic idea of minimax algorithm is synonymously related to the following optimization procedure. Max player tries as much as possible to increase the minimum value of the game, while Min tends to decrease its maximum value at node n as both players play towards optimality. The entire process can be formally described by the following extended Stockman formula (4) below:

International Journal of Game Theory and Technology (IJGTT) , Vol.1,No.1,June 2013

$$f(n) = \begin{cases} \max\{f(c) \mid c \text{ is a child node of } n\} - f(n), & \text{if } n \text{ is a min node} \\ \min\{f(c) \mid c \text{ is a child node of } n\} + f(n), & \text{if } n \text{ is a max node} \end{cases} \quad (5)$$

The minimax search equation combines with equation 1 to evolve the player where $x_{ai}$ and $y_{aj}$ are the average values of $x_i$ and $y_j$, respectively, and m is the number of pits on the Ayo board.

Furthermore a tournament was conducted between Minimax, Minimax-CBR and Awale (grandmaster) and the results are shown in Table 2 [16].The results furthermore show that CBR defeated all its opponents successfully.

Table 2.Case Based Reasoning

| MINIMAX(STD) | AWALE(STD) | MOVES(STD) | 0VERRIDES |
|---|---|---|---|
| 16.00(5.27) | 26.50(0.53) | 68.00(45.33) | NOT APPLICABLE |
| | | | |
| MINIMAX | MINIMAX-CBR | MOVES | OVERRIDES |
| 7.00(3.16) | 28.00(3.16) | 38.50(11.92) | 10.10(2.23) |
| | | | |
| MINIMAX-CBR | AWALE | MOVES | OVERRIDES |
| 25.50(0.53) | 15.00(1.05) | 42.70(2.31) | 24.00(2.11) |

## 3. Unsupervised machine Learning technique(UMLT)

This form of learning is best based on pattern recognition and clustering, it does not necessitate or need the correct results during training. Its unique characteristic is to find unreavealed patterns or hidden clusters in data sets which assist it in getting the right results. In can be used to cluster the input data in classes on the basis of their statistical properties only. There is significant clustering presence in unsupervised learning. Unsupervised learning refers to the problem of trying to find hidden structure in unlabelled data some of the examples include clustering (k-means, mixture models, hierarchical clustering)[20,21,22] and blind signal separation. The general technique used in unsupervised learning is described in Figure 2.The process can be grouped into 5 stages which are Training, Feature vector, Machine learning algorithm, Model and Better clustering classification[23].

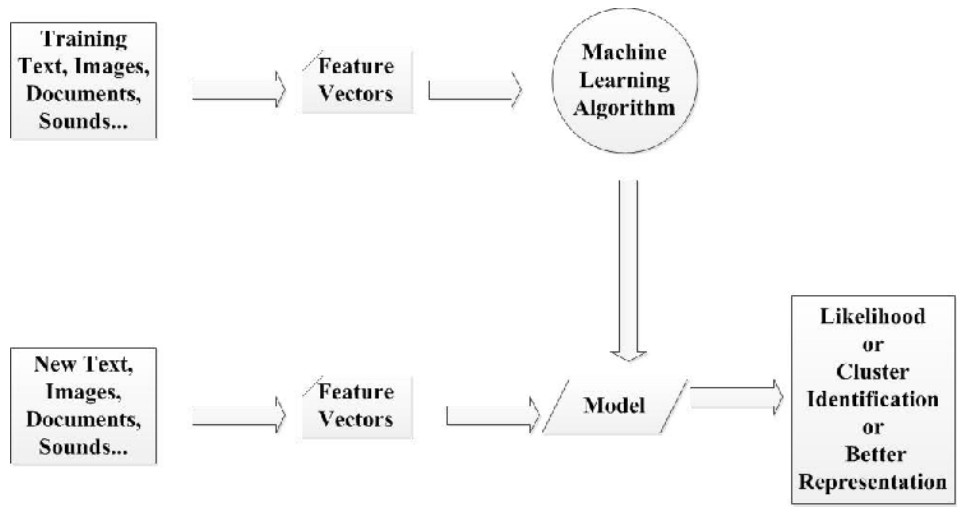

Figure 2.  Process of Unsupervised Machine Learning [23]

Most researches have been investiging supervised machine learning techniques and paying little attention to unsupervised learning techniques[8]but some researchers have been able to use these techniques to investigate, evolve or develop players to compete against the Awale shareware such as Probabilistic Distance Clustering(PDC)[1], Aggregate Mahalanobis Distance Function(ADMF)[24], Retrograde Analysis(RA)[25,26] and have all produced amazing results[2] but retrograde analysis is the only known unsupervised learning technique that has defeated Awale grandmaster conveniently.

Retrograde analysis technique is applicable to search spaces which can be completely enumerated within the memory of a computer system [27].RA first marks all end points such as checkmate, and then by making moves from the end positions works its way back to the positions farthest from the end positions, on the way determining the game-theoretical value of all positions in the search space. Retrograde analysis searches from bottom-up whereas other algorithms search from top-down such as Alpha-beta pruning, Breath-first and depth-first search.the advantage of RA is the fact that each position in the state space the optimal solution is determined [28], while other techniques which used top-down search technique only provided the optimal solution for a single starting point and the positions on the solution path.The study constructed a database using Godel numbers of the positions [29] which showed all the available positions in the database. Godel numbers were further modified to take unreachable positions into report. Each of the database entries stored scores between -48 and +48 and occupied 7 bits.

The database created by [26] was used to replay and analyse the games (Awale) at the computer Olympiad 2002 where the two strongest Awari/Awale programs competed against each other. The database performed very well and also overtook the playing strength of the world champion at the time, due to the fact that it stored scores rather than the best moves in the database [25]. The study realized that it was not always clear which move to take since there were multiple moves available that had very good scores.  RA[25,26] performed well against Awale shareware usings  889,063,398,406 positions enumerating all the possible states that can occur in the game. The database took 51 hours to construct on a 144 processor 1 GHz Pentium III cluster which was equipped with 72GB main memory and a 2 Gbs network. To ensure that the verification of the database the results were compared with results from two algorithms, different number of processors and the results obtained from other researchers and all these were done consistently. This technique has 2 major disadvantages (1) that it was too expensive to implement since Awari/Awale positions occurred in Billions and therefore such methods cannot be easily implemented on a small memory device like wireless handset [16] and . (2) The technique requires a huge amount of CPU time and internal memory which is caused by several expensive operations that are applied at each entry.

## 4. Analysis of supervised and unsupervised machine learning techniques used in evolving Awale player

Table 3 provides a performance analysis of popular machine learning techniques at various stages of the game, the table indicates what techniques were being used to evolve the agent/player, Table 3 shows the performance of the various agents(evolved players) and informs if the evolved player used minimax search technique or used endgame database or both. In the table the sign ( ) represents the fact that the process was successful or the technique or method was used while the sign($\times$) stands for unsuccessful or that technique was not used.

Table 3.performance of various Awale game players

| METHOD | MLT | | TECHNIQUE | | STAGES | | | |
|---|---|---|---|---|---|---|---|---|
| EVOLVED | SMLT | UMLT | MINIMAX | ENDGAME | INITIATION | BEGINNER | AMATEUR | GRANDMASTER |
| CBR |  | × |  |  |  |  |  |  |
| RAM-BPR |  | × |  |  |  |  |  | × |
| RAM-PRIORITY |  | × |  |  |  |  |  | × |
| RAM-CASING |  | × |  |  |  |  |  |  |
| ADMF | × |  |  |  |  |  |  | × |
| PDC | × |  |  |  |  |  |  | × |
| RA | × |  | × |  |  |  |  |  |
| CO-EVO |  | × |  | × |  |  |  | × |
| GA |  | × |  | × |  |  |  | × |
| NN |  | × | × | × |  |  |  | × |
| LDA |  | × |  |  |  |  |  | × |

All unsupervised machine learning techniques employed the use of databases which supported and enhanced their performance unlike some supervised techniques which did not employ the use of an endgame database, also there was no evolved player that was able to defeat the Awale shareware (grandmaster) conveniently without using the endgame databases to improve its performance. There is room for further improvement for the unsupervised machine learning techniques provided they improve their endgame database so as to enhance their performance against the Awale shareware.

## Conclusion

This has been an interesting study and the comparism of the various popular machine learning techniques in evolving Awale game player. Further studies and investigations will take an in-depth look at the various algorithms which have been used and what were the issues limiting their performance.